\documentclass[10pt,twocolumn,letterpaper]{article}

\usepackage[pagenumbers]{cvpr} 

\definecolor{cvprblue}{rgb}{0.21,0.49,0.74}
\usepackage[pagebackref,breaklinks,colorlinks,allcolors=cvprblue]{hyperref}

\usepackage{multirow}
\usepackage{amsmath}         
\usepackage{amssymb}         
\usepackage{algorithm}       
\usepackage{algpseudocode}   

\usepackage[T1]{fontenc} 

\definecolor{unsipervised}{RGB}{213, 232, 212}
\definecolor{supervised}{RGB}{248, 206, 204}
\definecolor{ourssuper}{RGB}{200, 200, 225}
\definecolor{best}{HTML}{fdf4bb}

\algnewcommand\Input{\item[\textbf{Input:}]}
\algnewcommand\Output{\item[\textbf{Output:}]}
\algnewcommand\Parameters{\item[\textbf{Parameters:}]}
\algnewcommand\Define{\item[\textbf{Define:}]}

\def\paperID{9994} 
\def\confName{CVPR}
\def\confYear{2026}

\title{Experts-Guided Unbalanced Optimal Transport for ISP Learning from Unpaired and/or Paired Data}

\author{Georgy Perevozchikov$^{1}$ \quad Nancy Mehta$^{1}$$^{*}$ \quad Egor Ershov$^{2,3,4}$ \quad Radu Timofte$^{1}$ \vspace{0.3em} \\
{\normalsize $^1$Computer Vision Lab, CAIDAS \& IFI, University of W\"urzburg} \quad
{\normalsize $^2$Institute for Information Transmission Problems, RAS} \quad \\
{\normalsize $^3$Moscow Institute of Physics and Technologies} \quad
{\normalsize $^4$Artificial Intelligence Research Institute} \\
{\tt\small \{georgii.perevozchikov,nancy.mehta,radu.timofte\}@uni-wuerzburg.de}
\quad
{\tt\small ershov@iitp.ru}
}

\begin{document}
\maketitle
\def\thefootnote{*}\footnotetext{Corresponding Author}
\begin{abstract}
Learned Image Signal Processing (ISP) pipelines offer powerful end-to-end performance but are critically dependent on large-scale paired raw-to-sRGB datasets.
This reliance on costly-to-acquire paired data remains a significant bottleneck.
To address this challenge, we introduce a novel, unsupervised training framework based on Optimal Transport capable of training arbitrary ISP architectures in both unpaired and paired modes.
We are the first to successfully apply Unbalanced Optimal Transport (UOT) for this complex, cross-domain translation task. 
Our UOT-based framework provides robustness to outliers in the target sRGB data, allowing it to discount atypical samples that would be prohibitively costly to map.
A key component of our framework is a novel ``committee of expert discriminators,'' a hybrid adversarial regularizer.
This committee guides the optimal transport mapping by providing specialized, targeted gradients to correct specific ISP failure modes, including color fidelity, structural artifacts, and frequency-domain realism.
To demonstrate the superiority of our approach, we retrained existing state-of-the-art ISP architectures using our paired and unpaired setups.
Our experiments show that while our framework, when trained in paired mode, exceeds the performance of the original paired methods across all metrics, our unpaired mode concurrently achieves quantitative and qualitative performance that rivals, and in some cases surpasses, the original paired-trained counterparts. The code and pre-trained models are available at: \url{https://github.com/gosha20777/EGUOT-ISP.git}.
 
\end{abstract}

\begin{figure}[t]
  \centering
  \includegraphics[width=1.0\linewidth]{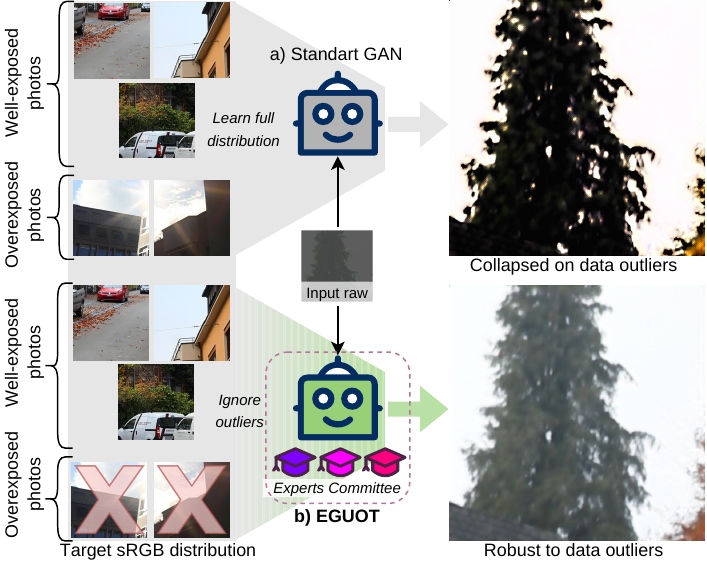}
    \vspace{-6mm}
   \caption{A conceptual overview of our framework's robustness to outliers. \textit{Top:} Standard GANs (RawFormer~\cite{perevozchikov2024rawformer}), when trained on unpaired datasets, are highly sensitive to data outliers (e.g., overexposed photos). This leads to training instability and a collapsed output. \textit{Bottom:} Our proposed EGUOT framework, leveraging Experts Guided Unbalanced Optimal Transport, is inherently robust to such data outliers and focuses on the high-quality core of the target distribution to produce perceptually better results.}
   \label{fig:teaser}
   \vspace{-4mm}
\end{figure}
    
\vspace{-1mm}
\section{Introduction}
\label{sec:introduction}
\vspace{-1mm}
Modern smartphone photography is a triumph of computational imaging, relying on a complex Image Signal Processor (ISP)~\cite{heide2014flexisp, karaimer2016software, delbracio2021mobile} to overcome the physical limitations of small sensors and compact optics~\cite{ignatov2017dslr, wronski2019handheld}. 
The ISP converts noisy, monochromatic, camera-specific raw sensor data into the perceptually-pleasing sRGB images users expect \cite{hasinoff2016burst, delbracio2021mobile}. While traditional hardware ISPs use fixed, hand-tuned modules \cite{brown2023color}, recent work has shifted to learned end-to-end pipelines \cite{zhang2021learning, xing2021invertible, ignatov2020aim}. These deep learning models offer superior performance and flexibility, bridging the quality gap between mobile phones and high-end DSLR cameras~\cite{ignatov2025learned, ignatov2020aim, ignatov2020replacing, ignatov2022microisp, ren2025ispformer}.

Despite their success, these state-of-the-art learned ISPs \cite{ignatov2025learned, ren2025ispformer, wirzberger2022lan, ignatov2020replacing, ignatov2020aim, ignatov2022microisp, ignatov2021learned, zhang2021learning} are critically dependent on large-scale, paired training datasets~\cite{shekhar2022transform, ignatov2020replacing, ignatov2025learned}. 
This requires a difficult and costly data acquisition process: capturing thousands of scenes simultaneously with a smartphone (raw) and a high-end reference DSLR (sRGB), followed by meticulous alignment and manual photo finishing~\cite{ignatov2020replacing, ershov2020cube++, ignatov2025learned}. 
This dependency forms a significant bottleneck, as the entire data collection and alignment effort must be repeated for every new camera sensor~\cite{nguyen2014raw, perevozchikov2024rawformer}.

The obvious solution is unpaired training to break this reliance on paired data. 
Recent attempts \cite{perevozchikov2024rawformer, arhire2025learned, nikonorov2025color} have explored this using standard Generative Adversarial Network (GAN) frameworks \cite{pang2021image, zhu2017unpaired, torbunov2023uvcgan2}, but these approaches often struggle with training stability and produce sub-optimal results~\cite{nikonorov2025color}. We assume that a core reason for this failure is the presence of distributional outliers (e.g., blurry, over/under-exposed images) within the raw-to-sRGB datasets~\cite{shekhar2022transform, ignatov2020replacing, ignatov2025learned}. Standard adversarial losses, including those based on classical Optimal Transport (OT)~\cite{korotin2022neural, gushchin2023entropic, kantorovitch1958translocation}, are highly sensitive to these outliers~\cite{choi2023generative}, as model performance is degraded by attempting to learn from such suboptimal data.

To overcome these challenges, we introduce \textbf{EGUOT} (\textbf{E}xperts-\textbf{G}uided \textbf{U}nbalanced \textbf{O}ptimal \textbf{T}ransport), a novel training framework robust to dataset outliers (Fig. \ref{fig:teaser}) that operates in \textit{paired} and \textit{unpaired} modes. Our framework is the first to apply Unbalanced Optimal Transport (UOT) \cite{pham2020unbalanced} for the raw-to-sRGB translation task. By relaxing classical OT constraints~\cite{kantorovitch1958translocation, korotin2022neural}, our method is inherently robust to outliers, discounting low-quality data to focus on the high-quality core of the target distribution. 
While UOT provides robustness~\cite{pham2020unbalanced, choi2023generative}, it is insufficient for ISP high-fidelity demands. We guide the transport mapping with a central component: a Committee of Expert Discriminators. This novel hybrid regularizer provides targeted, multi-modal gradients to correct specific ISP failure modes, such as semantic color inaccuracies, lack of structural detail, and unrealistic frequency artifacts.

We demonstrate the power and architecture-agnostic nature of our framework. 
We use our unsupervised method to retrain existing paired state-of-the-art supervised ISP backbones (e.g., MW-ISPNet \cite{ignatov2020aim}, Restormer \cite{zamir2022restormer}). 
Experiments confirm our EGUOT framework's flexibility. When trained without paired data, EGUOT achieves a new state-of-the-art for unpaired translation, handily beating prior work \cite{arhire2025learned, perevozchikov2024rawformer}. Its performance also proves competitive with, and at times better than, the original fully-paired models. Additionally, in a paired-data setting, our method establishes a new performance ceiling, uniformly improving upon the original models' results across all metrics.

\subsection*{Contribution}
\vspace{-2mm}
Our contributions can be summarized as follows:
\begin{itemize}
    \item We propose a novel framework for raw-to-sRGB translation, trainable in both \textit{unpaired} and \textit{paired} modes. To our knowledge, we are the first to successfully apply Unbalanced Optimal Transport (UOT)~\cite{choi2023generative} to this task, leveraging its robustness to dataset outliers, a critical flaw in prior adversarial approaches.

    \item We introduce a hybrid regularization method, the \textit{Committee of Expert Discriminators}, which guides the UOT mapping. This committee provides targeted, multi-modal perceptual gradients (color, structure, frequency), working in synergy with the UOT loss to ensure a high-fidelity image reconstruction rich in perceptual detail.

    \item We demonstrate that our \textit{architecture-agnostic framework} can train existing SOTA paired ISP backbones. In \textit{unpaired} mode, our models achieve performance competitive with, or even surpassing, their original, fully-paired results. In \textit{paired} mode, our framework surpasses these original results on standard benchmarks.
\end{itemize}

\section{Related Work}
\label{sec:related}

Our work is positioned at the intersection of three research domains: (1) learned ISP pipelines, (2) unpaired image-to-image translation, and (3) generative modeling with Optimal Transport.

\subsection{Learned Image Signal Processing}
The camera Image Signal Processor (ISP)~\cite{heide2014flexisp, karaimer2016software, delbracio2021mobile, chen2018learning, zamir2020cycleisp, xing2021invertible, Ignatov_2021_CVPR, zhang2021learning, jeong2022rawtobit, he2024enhancing} solves a complex image-to-image translation task, converting noisy, sensor-specific raw data into a plausible sRGB image \cite{hasinoff2016burst, delbracio2021mobile}. While traditional ISPs rely on a cascade of hand-tuned hardware modules \cite{brown2023color, heide2014flexisp, karaimer2016software, vsindelavr2013image, barron2017fast, herrmann2020learning, a2021beyond, jiang2022fast, conde2023perceptual}, this paradigm has recently been challenged by recent end-to-end deep learning approaches~\cite{zhang2021learning, xing2021invertible, ignatov2025learned, ignatov2020aim, ignatov2020replacing, ignatov2022microisp, ren2025ispformer, zheng2025evraw, mo2025diffuse}.


Foundational works (DeepISP \cite{schwartz2018deepisp}, PyNET \cite{ignatov2020replacing}) demonstrated that a U-Net-like network \cite{ronneberger2015u} could learn the entire ISP mapping and subsequent research improved their performance. MW-ISPNet \cite{ignatov2020aim} and AW-Net \cite{dai2020awnet} introduced Discrete Wavelet Transforms (DWT) to preserve high-frequency details. LAN-ISP \cite{wirzberger2022lan} and \cite{ignatov2022microisp, ignatov2025learned, zhang2021learning} integrated attention mechanisms for improved spatial processing. More recently, Transformer-based~\cite{dosovitskiy2020image} backbones like Restormer \cite{zamir2022restormer}, ISPFormer \cite{ren2025ispformer}, and \cite{brateanu2025lyt, zheng2025evraw, hong2025srrsr, he2024enhancing} were adapted for this task, leveraging powerful long-range dependency modeling.

Concurrently, specialized networks were developed for specific ISP sub-problems~\cite{seizinger2023efficient, cai2023retinexformer, afifi2020deep, afifi2021learning, conde2024instructir, conde2023perceptual}, including raw-to-raw translation \cite{perevozchikov2024rawformer, afifi2021semi} for generalizing ISPs to new camera sensors, and color mapping (e.g., cmKAN \cite{nikonorov2025color}, NiLUT~\cite{conde2024nilut}, SepLUT~\cite{yang2022seplut}, and~\cite{li2024sirlut, ke2023neural, chang2015palette, afifi2021histogan, ding2024regional, liu20234d, brateanu2025enhancing}) for high-fidelity color matching.

Despite architectural diversity and strong performance, these state-of-the-art methods are all trained in a fully paired manner, fundamentally depending on large-scale, paired raw-to-sRGB datasets \cite{shekhar2022transform, ignatov2020replacing, ershov2020cube++, ignatov2025learned, banic2024ntire}. This data is notoriously difficult and costly to acquire, requiring meticulous spatial and temporal alignment between a source (e.g., smartphone) and a target (e.g., DSLR) camera. This paired data bottleneck \cite{perevozchikov2024rawformer, nikonorov2025color, afifi2021semi, arhire2025learned} is the field's primary limitation, as the entire data collection process must be repeated for every new camera sensor. This limitation motivates our work in developing a robust unpaired framework.

\subsection{Unpaired Image-to-Image Translation}
The most common approach for unpaired image-to-image translation is the Generative Adversarial Network (GAN)~\cite{pang2021image}. CycleGAN \cite{zhu2017unpaired} introduced cycle-consistency, using a pair of generators and discriminators to enforce that a translated image can map back to its original domain. This inspired a wide range of methods based on shared latent spaces (e.g., DualGAN~\cite{yi2017dualgan}, UNIT~\cite{liu2017unsupervised}, STARGAN~\cite{choi2018stargan}, SEAN~\cite{zhu2020sean}), adaptive normalization (e.g., U-GAT-IT \cite{kim2019u}, UVCGANv2~\cite{torbunov2023uvcgan2}), contrastive learning (e.g., CUT \cite{park2020contrastive}), and recent~\cite{torbunov2023rethinking, choi2020stargan, cheng2020sequential, zhao2020unpaired}.

Last years, this unpaired approach was explored for the raw-to-sRGB task \cite{arhire2025learned}, proposing the usage of a standard GAN~\cite{pang2021image} with a multi-term loss, including pre-trained networks (VGG~\cite{ledig2017photo}, LPIPS~\cite{zhang2018unreasonable}) for content preservation. While pioneering, this approach relies on standard adversarial losses~\cite{pang2021image}, not designed to handle distributional outliers in real-world datasets~\cite{nikonorov2025color, choi2023generative}. We hypothesize that standard GAN losses are negatively impacted by these outliers, leading to sub-optimal results quality, thus necessitating a more robust adversarial framework.

\subsection{Optimal Transport in Generative Modeling}
\label{sec:related_ot}

Optimal Transport (OT)~\cite{kantorovitch1958translocation} offers a powerful theoretically-grounded alternative to traditional GAN losses. While many methods use the OT cost as a loss function (e.g., Wasserstein GANs~\cite{arjovsky2017wasserstein, korotin2019wasserstein}), recent works have shown the OT plan itself can serve as a generative model~\cite{korotin2022kantorovich, korotin2022neural, gushchin2023entropic, makkuva2020optimal, pooladian2024neural, geuter2025universal, gazdieva2025optimal, vasa2025context}.

However, a well-known limitation of classical OT is its high sensitivity to outliers~\cite{choi2023generative, korotin2022neural, pham2020unbalanced}. The standard OT problem requires transporting every point in the source distribution to the target distribution. This means a few outlier samples can unduly influence the data~\cite{choi2023generative}, drastically altering the transport map and degrading the final result.

To solve this, the theory of \textit{Unbalanced Optimal Transport (UOT)} was proposed~\cite{pham2020unbalanced}, relaxing the hard marginal constraints of classical OT~\cite{kantorovitch1958translocation, korotin2022neural}. This relaxation permits partial transport, which, in practical terms, means the model can learn to \textit{ignore or discard outliers} rather than being forced to match them.

While the UOT theory was powerful, a stable, practical algorithm was needed. The UOTM paper \cite{choi2023generative} introduced a novel, stable training framework based on the semi-dual formulation of UOT \cite{pham2020unbalanced}. The authors demonstrated that UOTM is significantly more robust to outliers and converges faster than classical OT models \cite{kantorovitch1958translocation, korotin2022neural}.

Our work is the first to connect these fields. We are the first to apply the Experts Guided UOT framework to the paired and unpaired ISP problem. Thus, we leverage the robustness of UOT \cite{choi2023generative} to solve a task where prior methods struggle precisely because of the outlier sensitivity that UOT was designed to fix \cite{choi2023generative}.
\section{Method}
\label{sec:method}

\begin{figure*}[t]
  \centering
  \includegraphics[width=1.0\linewidth]{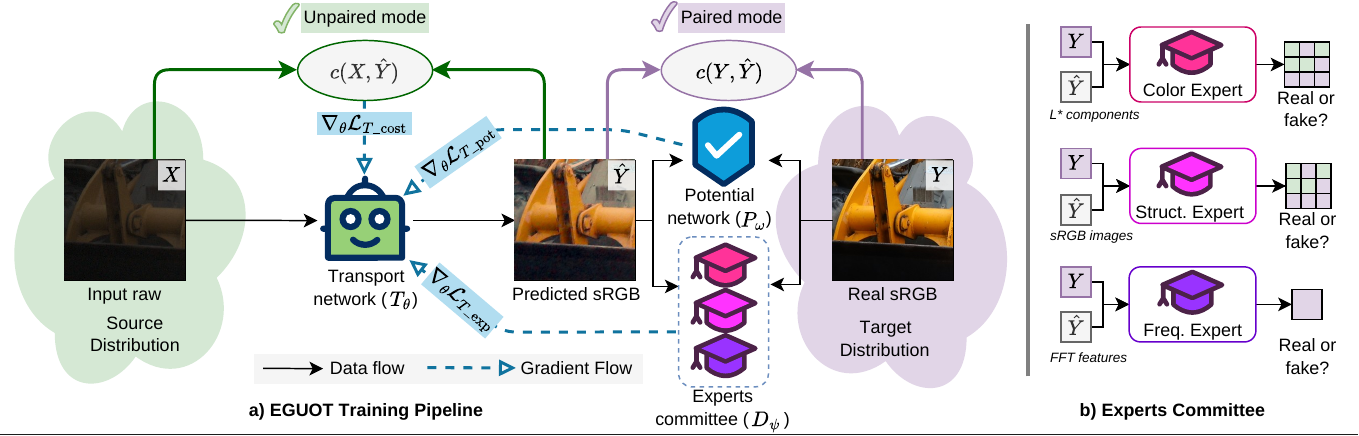}
  \vspace{-4mm}  
   \caption{\textbf{An overview of our EGUOT training framework.} (a) \textit{The main pipeline}, where the generator ($T_\theta$) is trained by three parallel objectives: (1) a content-preserving cost function ($c(\cdot,\cdot)$) for paired and unpaired mode, (2) a robust Potential network ($P_\omega$), and (3) an Experts committee ($D_\psi$). (b) \textit{The detailed breakdown of the Experts committee}, showing its three specialized discriminators for \textit{Color}, \textit{Structure}, and \textit{Frequency}, each analyzing different features of the real ($Y$) and predicted ($\hat{Y}$) sRGB images.}
   \label{fig:overview}
   \vspace{-4mm}
\end{figure*}

Our goal is to train a raw-to-sRGB mapping network $T_\theta$ for \textit{paired} and \textit{unpaired} settings, which we achieve with EGUOT, a flexible training framework (Fig. \ref{fig:overview}). The framework consists of a generator $T_\theta$ trained by a novel hybrid loss system. In unpaired mode, this system uses a cost function $c(x, T_\theta(x))$ for robust content mapping and a Committee of Expert Discriminators $D_\psi$ for perceptual fidelity. In paired mode, the system has a direct pixel loss $c(y, T_\theta(x))$, retaining the expert committee.

\subsection{Preliminaries: UOT for Image Translation}
\label{sec:preliminaries}

Inspired by UOTM~\cite{choi2023generative}, which learns a transport $T_\theta$ and potential $P_\omega$ network via a minimax problem, we utilize its core components:
\begin{itemize}
    \item $T_\theta$: The transport network that learns the transport map.
    \item $P_\omega$: The potential network that acts as the discriminator.
    \item $c(x, y)$: A function measuring the $x$ to $y$ mapping cost.
    \item $\Phi_1, \Phi_2$: A non-decreasing, convex function pair (e.g., $e^x$) applying a soft penalty that relaxes transport constraints, enabling outlier robustness.
\end{itemize}

The objective for the potential network $P_\omega$, augmented with a stabilizing $R_1$ gradient penalty \cite{roth2017stabilizing}, is:
\begin{equation}\label{eq:potential_obj}
\begin{aligned}
\mathcal{L}_P = & \mathbb{E}_{x \sim \mathbb{X}}[\Phi_1(-c(x, T_\theta(x)) + P_\omega(T_\theta(x)))] \\
& + \mathbb{E}_{y \sim \mathbb{Y}}[\Phi_2(-P_\omega(y))] + \frac{\gamma}{2} \mathbb{E}_{y \sim \mathbb{Y}} [\|\nabla_{y} P_\omega(y)\|_2^2]
\end{aligned}
\end{equation}

The objective for the transport network $T_\theta$ is:
\begin{equation}\label{eq:transport_obj}
\mathcal{L}_T = \mathbb{E}_{x \sim \mathbb{X}} [c(x, T_\theta(x)) - P_\omega(T_\theta(x))]
\end{equation}

UOTM~\cite{choi2023generative} application to the raw-to-sRGB task requires two considerations: (1) a valid cost function $c(\cdot, \cdot)$ must be defined for the paired/unpaired case, as raw and sRGB domains are incompatible; and (2) the basic transport loss (Eq.~\ref{eq:transport_obj}) is insufficient alone for the high-fidelity translation quality. Our EGUOT framework addresses both.

\subsection{The EGUOT Framework}
We propose the EGUOT framework to solve these two challenges. First, we define the cost functions to utilize Eq. (\ref{eq:potential_obj}) and (\ref{eq:transport_obj}). Second, we introduce a hybrid loss function that regularizes the objective (\ref{eq:potential_obj}) with a committee of expert discriminators.

\subsubsection{Cost Functions}

We define the transport cost $c(\cdot, \cdot)$ for paired and unpaired training.

\textbf{Unpaired Setting:} The cost $c(x, T_\theta(x))$ is computed between the input $x$ (4-channel raw) and output $T_\theta(x)$ (3-channel sRGB). To resolve this domain incompatibility, we use a fixed pre-processing function $P$ (e.g., basic demosaicing) mapping the raw image $x \in \mathbb{X}$ to a simple RGB image $P(x) \in \mathcal{X}$. The unpaired cost is the $L_2$ distance:

$$
c(x, T_\theta(x)) = \tau \|T_\theta(x) - P(x)\|_2^2
$$

\textbf{Paired Setting:} With access to the ground-truth sRGB image $y$, we use $L_1$ norm as a stronger paired cost.

\subsubsection{Experts-Guided Hybrid Loss}
The UOTM~\cite{choi2023generative} objective, while robust, is insufficient to enforce the high-quality raw-to-sRGB mapping.

To address this, we introduce a committee of expert discriminators $D_\psi = \{D_i\}_{i=1}^M$, which runs in parallel to the Potential Network $P_\omega$ providing targeted adversarial gradients. We use a Hinge loss $h(t) = \max(0, 1+t)$ for this committee.

The expert discriminators $D_\psi$ are trained to distinguish real sRGB images from fake ones by minimizing $\mathcal{L}_{D}$:
$$
\mathcal{L}_{D} = \sum_{D_i \in D_\psi} \left( \mathbb{E}_{y \sim \mathbb{Y}}[h(-D_i(y))] + \mathbb{E}_{x \sim \mathbb{X}}[h(D_i(T_\theta(x)))] \right)
$$
The generator $T_\theta$ is then simultaneously trained to fool this committee by minimizing the expert generator loss $\mathcal{L}_{T\_\text{exp}}$:
$$
\mathcal{L}_{T\_\text{exp}} = \sum_{D_i \in D_\psi} \mathbb{E}_{x \sim \mathbb{X}}[-D_i(T_\theta(x))]
$$
The final, total loss for our generator is:
$$
\mathcal{L}_{T\_\text{total}} = \mathcal{L}_T + \lambda \cdot \mathcal{L}_{T\_\text{exp}}
$$

\subsection{Network Architectures}
We selected SOTA components for our approach.

\textbf{Transport Network ($T_\theta$):} A learnable ISP pipeline for the complete raw-to-sRGB mapping. As our framework is architecture-agnostic, we adopt multiple SOTA ISP backbones (e.g., \cite{nikonorov2025color, ignatov2025learned}) to serve as $T_\theta$ in our experiments.

\textbf{Potential Network ($P_\omega$):} This network serves as the potential function~\cite{korotin2022neural}. We use a ConvNeXt~\cite{liu2022convnet} backbone, whose features are passed through an MLP head to output a single scalar.

\textbf{Experts Committee ($D_\psi$):} The committee is comprised of three specialist networks:
\begin{itemize}
    \item \textbf{Color Expert:} A frozen ConvNeXt \cite{liu2022convnet} encoder from a pre-trained colorization U-Net \cite{ronneberger2015u} that judges semantic color plausibility in the L* channel.
    \item \textbf{Structure Expert:} A multi-scale PatchGAN discriminator~\cite{perevozchikov2024rawformer} operating on the sRGB image to enforce local texture fidelity.
    \item \textbf{Frequency Expert:} A lightweight CNN operating on the 2D log-magnitude FFT spectrum of the grayscale image to penalize high-frequency artifacts.
\end{itemize}

\subsection{Training Algorithm}
\vspace{-2mm}
We train our framework with an alternating optimization scheme. We use a discriminator-heavy $N:1$ training ratio: both the Potential Network $P_\omega$ and Experts Committee $D_\psi$ are updated $N$ steps for every single Transport Network $T_\theta$ update. The complete EGUOT training procedure is in Algorithm \ref{alg:uot-ex}.

\section{Experiments}
\label{sec:experiments}
\vspace{-2mm}
To validate our EGUOT framework, we conduct a comprehensive set of experiments. We demonstrate that (1) our framework can train existing supervised ISP backbones in paired and unpaired modes to outperform (in the paired case) or achieve performance competitive with (in the unpaired case) their original settings; (2) our framework sets a new state-of-the-art for unpaired raw-to-sRGB translation; (3) the core components of our method---the UOT objective and the expert committee---are both essential for its success.

\begin{algorithm}[h]
\caption{ISP with EGUOT}
\label{alg:uot-ex}
\begin{algorithmic}[1]
\Input The source $\mathbb{X}$ and the target $\mathbb{Y}$ distributions; transport $T_\theta: \mathbb{R}^X \to \mathbb{R}^Y$; potential $P_\omega: \mathbb{R}^Y \to \mathbb{R}$; experts $D_\psi = \{D_i: \mathbb{R}^Y \to \mathbb{R}\}_{i=1}^M$; experts loss $h: \mathbb{R} \to \mathbb{R}$; cost $c: \mathbb{Y} \times \mathbb{Y} \to \mathbb{R}$ for \textit{paired} or $c: \mathbb{X} \times \mathbb{Y} \to \mathbb{R}$ for \textit{unpaired} mode; non-decreasing function pair ($\Phi_1$, $\Phi_2$); number of inner iterations $N$; hyperparameters $\gamma, \lambda$.
\Output learned transport network $T_\theta$.

\Repeat \Comment{Discriminator/Potential training loop}
    \For{$k = 1, \dots, N$}
        \State Sample batch $X \sim \mathbb{X}$, $Y \sim \mathbb{Y}$;
        
        \State {\scriptsize $\begin{aligned}[t] \mathcal{L}_P \gets & \frac{1}{|X|}\sum_{x \in X} \Phi_1\left(-c(\cdot, T_\theta(x)) + P_\omega(T_\theta(x))\right) \\
                        & + \frac{1}{|Y|}\sum_{y \in Y} \Phi_2(-P_\omega(y)) \\
                        & + \frac{\gamma}{2} \cdot \frac{1}{|Y|}\sum_{y \in Y} \|\nabla_{y} P_\omega(y)\|_2^2; \end{aligned}$}
        \State Update $\omega$ by using $\nabla_\omega \mathcal{L}_P$;
        
        \State $\begin{aligned}[t] \mathcal{L}_{D} \gets \sum_{D_i \in D_\psi} \bigg( & \frac{1}{|Y|}\sum_{y \in Y} h(-D_i(y)) \\
                                & + \frac{1}{|X|}\sum_{x \in X} h(D_i(T_\theta(x))) \bigg); \end{aligned}$
        \State Update $\psi$ by using $\nabla_\psi \mathcal{L}_{D}$;
    \EndFor
    \State Sample batch $X \sim \mathbb{X}$; \Comment{Transport loss}
    \State $\mathcal{L}_{T} \gets \frac{1}{|X|}\sum_{x \in X} \left(\underbrace{c(\cdot, T_\theta(x))}_{\mathcal{L}_{T\_\text{cost}}} - \underbrace{P_\omega(T_\theta(x))}_{\mathcal{L}_{T\_\text{pot}}}\right) + \lambda \cdot \underbrace{\sum_{D_i \in D_\psi} \frac{1}{|X|}\sum_{x \in X}[-D_i(T_\theta(x))]}_{\mathcal{L}_{T\_\text{exp}}}$;
    \State Update $\theta$ by using $\nabla_\theta \mathcal{L}_{T}$;

\Until{\textit{not converged};}
\end{algorithmic}
\end{algorithm}
\vspace{-4mm}

\subsection{Datasets}

\paragraph{Zurich raw-to-sRGB (ZRR) dataset~\cite{ignatov2020replacing}} This dataset consists of 48,043 paired raw (Huawei P20) and sRGB (Canon 5D Mark IV) image patches (448$\times$448). We follow the official dataset splits for training and testing.
\vspace{-4mm}

\paragraph{ISPIW dataset~\cite{shekhar2022transform}} This dataset comprises raw images from a Huawei Mate 30 Pro and corresponding sRGB references from a Canon 5D Mark IV. We use all 192 full-resolution images, which are split into 5,724 patches, using a 70\%/20\%/10\% split for training, validation, and testing.
\vspace{-4mm}

\paragraph{Mobile AI dataset~\cite{ignatov2020aim}} This dataset was acquired using a Sony IMX586 Quad Bayer sensor and a Fujifilm GFX100 DSLR. The dataset contains 95K cropped 256$\times$256 image pairs. We again follow the standard 70\%/20\%/10\% split.

\subsection{Training Details}

All experiments were conducted on a single NVIDIA GeForce RTX 4090 GPU. All networks (Transport $T_\theta$, Potential $P_\omega$, and Expert Committee $D_\psi$) were trained using the Adam optimizer with $\beta_1=0.5$ and $\beta_2=0.999$. We used a Cosine Annealing scheduler for all learning rates. The learning rates for the transport, potential, and expert networks were set to $lr_{\text{transport}} = 1e-4$, $lr_{\text{potential}} = 2e-4$, and $lr_{\text{experts}} = 2e-4$, respectively. The $R_1$ gradient penalty coefficient was $\gamma = 1.5$. The UOT content cost weight was $\tau = 1e-3$, and the expert committee weight was $\lambda = 1.0$. All models were trained for 300 epochs. For the final model weights, we apply Stochastic Weight Averaging (SWA) for the last 10 epochs. For all unpaired training tasks, the source and target datasets were shuffled independently.

\subsection{Results}

\begin{figure*}[t!]
  \centering
  \includegraphics[width=1.0\linewidth]{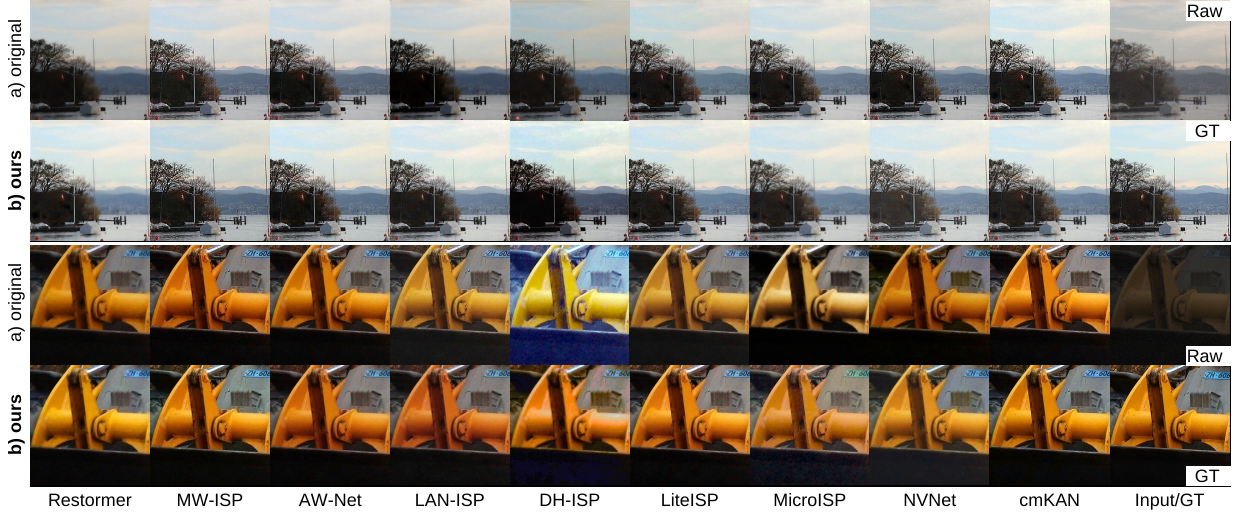}

   \vspace{-2mm}
   \caption{Our framework's unpaired mode achieves the paired performance across diverse architectures on the Zurich raw-to-sRGB dataset~\cite{ignatov2020replacing}. The top row (a) shows the results of SOTA ISP backbones (e.g., Restormer~\cite{perevozchikov2024rawformer}, cmKAN~\cite{nikonorov2025color}) trained with their \textit{original paired} settings. The bottom row (b) --- results of the same backbones trained in \textit{unpaired mode}, highlighting the effectiveness of our method. Best viewed in the electronic version.}
   \label{fig:zrr-isp}
   \vspace{-3mm}
\end{figure*}

Our primary results demonstrate EGUOT's performance and flexibility. First, EGUOT can train diverse SOTA ISP architectures in an unpaired manner, achieving competitive performance, while our paired mode achieves superior performance. Finally, our framework sets a new state-of-the-art for unpaired raw-to-sRGB translation. \textit{See supp. materials for additional results.}

\subsubsection*{ZRR Results}

\begin{table*}[h]
  \centering
    \caption{Results on the Zurich raw-to-sRGB dataset~\cite{ignatov2020replacing}, comparing SOTA backbones retrained with our EGUOT framework against their \textit{original paired-data results}. Our \textit{paired} mode consistently outperforms the original, while our \textit{unpaired} mode achieves competitive performance. Better results are highlighted in yellow. \label{tab:zrr-isp}}
    \vspace{-2mm}
  \resizebox{1.\linewidth}{!}
  {
  \begin{tabular}{@{}l|cccc|cccc|cccc|ccc@{}}
    \toprule
    Backbone & PSNR $\uparrow$ & SSIM $\uparrow$ & $\Delta E$ $\downarrow$ & LPIPS $\downarrow$ & PSNR $\uparrow$ & SSIM $\uparrow$ & $\Delta E$ $\downarrow$ & LPIPS $\downarrow$ & PSNR $\uparrow$ & SSIM $\uparrow$ & $\Delta E$ $\downarrow$ & LPIPS $\downarrow$ & \#Params & FLOPs & Architecture Type\\
    & \multicolumn{4}{c|}{\colorbox{supervised}{Paired Mode (original)}} & \multicolumn{4}{c|}{\colorbox{ourssuper}{Paired Mode (ours)}} & \multicolumn{4}{c|}{\colorbox{unsipervised}{Unpaired Mode (ours)}}\\
    \midrule
    Restormer~\cite{zamir2022restormer}     & 21.51 & 0.83 & 9.71 & 0.19& \colorbox{best}{21.73} & \colorbox{best}{0.84} & \colorbox{best}{9.42} & \colorbox{best}{0.18} & \colorbox{best}{21.69} & \colorbox{best}{0.84} & 9.93 & 0.20 & 5.03M & 583G & U-Net-like \\
    MW-ISP~\cite{ignatov2020aim}     & 21.88 & 0.82 & 10.33 & 0.21& \colorbox{best}{22.15} & \colorbox{best}{0.84} & \colorbox{best}{9.61} & \colorbox{best}{0.20} & 21.62 & \colorbox{best}{0.83} & \colorbox{best}{9.97} & \colorbox{best}{0.20} & 29.22M & 3.6T & U-Net-like \\
    AW-Net~\cite{dai2020awnet}     & 21.58 & 0.75 & 11.02 & 0.19 & \colorbox{best}{22.02} & \colorbox{best}{0.77} & \colorbox{best}{9.98} & \colorbox{best}{0.18} & \colorbox{best}{21.71} & \colorbox{best}{0.76} & \colorbox{best}{10.17} & 0.19 & 54.83M & 3.1T & U-Net-like \\
    LAN-ISP~\cite{wirzberger2022lan} & 20.76 & 0.81 & 10.41 & 0.22 & \colorbox{best}{21.19} & 0.81 & \colorbox{best}{9.36} & \colorbox{best}{0.21} & 20.50 & 0.80 & \colorbox{best}{9.88} & 0.22 & 49.2K & 56.9G & U-Net-like \\
    LiteISP~\cite{zhang2021learning} & 22.18 & 0.83 & 10.28 & 0.19 & \colorbox{best}{22.29} & \colorbox{best}{0.84} & \colorbox{best}{9.79} & 0.19 & 21.43 & 0.83 & \colorbox{best}{10.04} & 0.21 & 9.04M & 174G & U-Net-like \\
    DH-ISP~\cite{ignatov2021learned} & 19.68 & 0.72 & 11.46 & 0.26 & \colorbox{best}{19.88} & \colorbox{best}{0.74} & \colorbox{best}{10.99} & \colorbox{best}{0.24} & \colorbox{best}{19.69} & \colorbox{best}{0.74} & \colorbox{best}{11.25} & \colorbox{best}{0.25} & 3.1K & 31.7G & Forward CNN \\
    MicroISP~\cite{ignatov2022microisp} & 20.30 & 0.78 & 11.14 & 0.24 & \colorbox{best}{20.61} & \colorbox{best}{0.79} & \colorbox{best}{11.08} & 0.24 & 19.81 & \colorbox{best}{0.79} & 11.65 & 0.24 & 105K & 37G & Forward CNN \\
    NVNet~\cite{ignatov2025learned} & 20.95 & 0.80 & 10.96 & 0.22 & \colorbox{best}{21.13} & 0.80 & \colorbox{best}{10.05} & \colorbox{best}{0.21} & 20.15 & 0.78 & \colorbox{best}{10.31} & 0.23 & 3.5K & 31.8G & ResNet-like \\
    cmKAN~\cite{nikonorov2025color} & 24.41 & 0.85 & 7.27 & 0.17 & \colorbox{best}{24.63} & \colorbox{best}{0.86} & \colorbox{best}{7.10} & 0.17 &\colorbox{best}{24.46} & 0.84 & 8.31 & 0.18 & 76.4K & 40G & Hyper-network \\
    \bottomrule
  \end{tabular}
  }  \vspace{-2mm}
\end{table*}

We begin our analysis with the ZRR dataset. Table \ref{tab:zrr-isp} and Figure \ref{fig:zrr-isp} show the results of applying our framework (paired and unpaired) to nine SOTA ISP backbones. The results are compelling: (1) \textit{our paired} mode consistently surpasses the original paired-trained models across all metrics and architectures, showing the advantage of our expert-guided regularization; (2) \textit{our unpaired} mode consistently rivals its paired counterparts. In particular, our EGUOT-trained models even \textit{surpass} the original results in several key metrics, such as SSIM for Restormer~\cite{zamir2022restormer} and MW-ISP~\cite{ignatov2020aim}, and $\Delta E$ for LAN-ISP~\cite{wirzberger2022lan} and AW-Net~\cite{dai2020awnet}. This demonstrates our framework is truly architecture-agnostic and can effectively replace the need for paired data.

\begin{figure}[h!]
  \centering
  \includegraphics[width=1.0\linewidth]{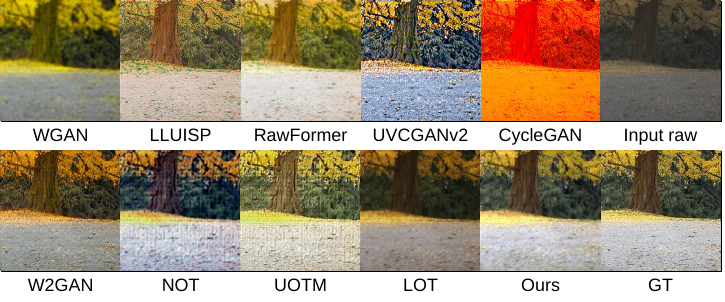}
    \vspace{-1mm}
   \caption{Quantitative results of \textit{unpaired frameworks} on the Zurich raw-to-sRGB dataset~\cite{ignatov2020replacing}. We compare against modern GAN-based (e.g., RawFormer~\cite{perevozchikov2024rawformer}, UVCGANv2~\cite{torbunov2023rethinking}), other OT-based (e.g., NOT~\cite{korotin2022neural}, UOTM~\cite{choi2023generative}), and recent unpaired ISP methods (LLUISP~\cite{arhire2025learned}). Our framework sets a \textit{new state-of-the-art performance across all metrics}. Best viewed in the electronic version.}
   \label{fig:zrr-gans}
   \vspace{-4mm}
\end{figure}

\begin{table}[t]
  \centering
    \caption{Results of various unapired image-to-image translation approaches on the Zurich raw-to-sRGB dataset~\cite{ignatov2020replacing}. We retrain all methods on a fixed Restormer backbone for a fair comparison. Our EGUOT framework achieves the best performance. \label{tab:zrr-unpaired}}
  \label{tab:cm-ours}
  \resizebox{0.95\linewidth}{!}
  {
  \begin{tabular}{@{}l|cccc|c@{}}
    \toprule
    Method & PSNR $\uparrow$ & SSIM $\uparrow$ & $\Delta E$ $\downarrow$  & LPIPS $\downarrow$ & Type\\
    \midrule
    CycleGAN~\cite{zhu2017unpaired} & 7.73 & 0.32 & 27.8 & 0.82 & GAN-based\\
    UVCGANv2~\cite{torbunov2023uvcgan2}& 16.25 & 0.61 & 12.2 & 0.49 & GAN-based\\
    RawFormer~\cite{perevozchikov2024rawformer}& 17.09 & 0.68 & 11.9 & 0.45 & GAN-based\\
    WGAN~\cite{arjovsky2017wasserstein} & 14.13 & 0.52 & 14.3 & 0.67 & GAN-based\\
    W2GAN~\cite{korotin2019wasserstein} & 15.62 & 0.58 & 13.9 & 0.60 & GAN-based\\
    LLUISP~\cite{arhire2025learned} & 19.07 & 0.72 & 11.7 & 0.32 & GAN-based\\
    NOT~\cite{korotin2022neural} & 16.03 & 0.60 & 12.6 & 0.54 & OT-based\\
    UOTM~\cite{choi2023generative} & 18.96 & 0.71 & 11.0 & 0.31 & OT-based\\
    LOT~\cite{pooladian2024neural} & 17.77 & 0.68 & 12.2 & 0.39 & OT-based\\
    Ours & \colorbox{best}{21.69} &  \colorbox{best}{0.84} &  \colorbox{best}{9.93} &  \colorbox{best}{0.20} &  OT-based\\
    \bottomrule
  \end{tabular} 
  }
\vspace{-4mm}
\end{table}

Next, we establish a new state-of-the-art for unpaired translation on this dataset. Table \ref{tab:zrr-unpaired} compares unpaired training frameworks using a fixed Restormer backbone. Our EGUOT framework significantly outperforms all other methods. It surpasses standard GANs (CycleGAN~\cite{zhu2017unpaired}, WGAN~\cite{arjovsky2017wasserstein}) by a massive margin and clearly outperforms recent specialized methods like RawFormer~\cite{perevozchikov2024rawformer} and LLUISP~\cite{arhire2025learned}. Crucially, EGUOT also outperforms other OT-based methods (NOT~\cite{korotin2022neural}, UOTM~\cite{choi2023generative}), demonstrating our novel experts-guided hybrid loss is critical for achieving high-fidelity results. Figure \ref{fig:zrr-gans} qualitatively shows our method's superior color and texture fidelity.

\subsubsection*{ISPIW Results}

\begin{figure}[t]
  \centering
  \includegraphics[width=1.0\linewidth]{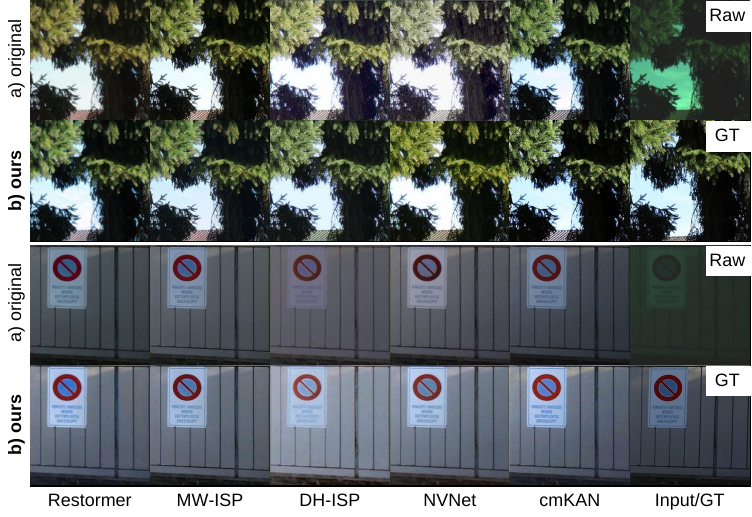}

   \caption{Quantitative results on the on the ISPIW raw-to-sRGB dataset~\cite{shekhar2022transform}, demonstrating the architecture-agnostic nature of our EGUOT framework. We apply (a) the original, \textit{paired} training and (b) our proposed \textit{unpaired} training to a wide range of SOTA ISP backbones. Our unpaired mode achieves performance that is highly competitive with the original settings. Best viewed in the electronic version.}
   \label{fig:ispiw-isp}
\end{figure}

\begin{table}[h]
  \centering
    \caption{Results on the ISPIW raw-to-sRGB dataset~\cite{shekhar2022transform}, comparing SOTA backbones retrained with our framework against their \textit{original paired-data results}. Our \textit{unpaired} mode achieves competitive performance. Better results are highlighted in yellow.\label{tab:ispiw-isp}}
  \resizebox{1.\linewidth}{!}
  {
  \begin{tabular}{@{}l|cccc|cccc@{}}
    \toprule
    Backbone & PSNR $\uparrow$ & SSIM $\uparrow$ & $\Delta E$ $\downarrow$  & LPIPS $\downarrow$ & PSNR $\uparrow$ & SSIM $\uparrow$ & $\Delta E$ $\downarrow$  & LPIPS $\downarrow$\\
    & \multicolumn{4}{c|}{\colorbox{supervised}{Paired Mode (original)}} & \multicolumn{4}{c}{\colorbox{unsipervised}{Unpaired Mode (ours)}}\\
    \midrule
    Restormer~\cite{zamir2022restormer}                & 20.93 & 0.79 & 6.85 & 0.14 & 20.81 & 0.79 & 6.99 & 0.14 \\
    MW-ISP~\cite{ignatov2020aim}     & 21.90 & 0.81 & 7.03 & 0.11 & 21.81 & 0.81 & \colorbox{best}{6.99} & 0.11 \\
    AW-Net~\cite{dai2020awnet}                   & 21.75 & 0.81 & 6.99 & 0.12 & \colorbox{best}{21.93} & 0.81 & 7.30 & \colorbox{best}{0.11} \\
    LAN-ISP~\cite{wirzberger2022lan}     & 22.09 & 0.81 & 6.97 & 0.11 & 21.95 & 0.81 & \colorbox{best}{6.42} & 0.11 \\
    LiteISP~\cite{zhang2021learning}     & 22.14 & 0.81 & 6.31 & 0.11 & 22.01 & 0.80 & 6.88 & 0.12 \\
    DH-ISP~\cite{ignatov2021learned}     & 19.92 & 0.76 & 7.89 & 0.17 & 19.59 & \colorbox{best}{0.77} & \colorbox{best}{7.40} & 0.17 \\
    MicroISP~\cite{ignatov2022microisp}     & 20.70 & 0.77 & 6.92 & 0.15 & \colorbox{best}{20.79} & 0.77 & \colorbox{best}{6.33} & 0.15 \\
    NVNet~\cite{ignatov2025learned}     & 23.80 & 0.82 & 5.81 & 0.10 & 23.39 & 0.82 & 6.08 & 0.10 \\
    cmKAN~\cite{nikonorov2025color}     & 24.22 & 0.83 & 5.29 & 0.09 & 24.21 & 0.83 & 5.73 & 0.09 \\
    \bottomrule
  \end{tabular}
  }
\end{table}

We confirm these findings with the ISPIW dataset. As shown in Table \ref{tab:ispiw-isp} and Figure \ref{fig:ispiw-isp}, our unpaired training again achieves performance that is statistically on-par with the original paired training for nearly all architectures. For several models (e.g., AW-Net~\cite{dai2020awnet}, MicroISP~\cite{ignatov2022microisp}), our unpaired method even achieves a higher PSNR or lower $\Delta E$ than the paired-data baseline, further highlighting the framework's effectiveness.

\subsubsection*{Mobile AI Results}

\begin{figure}[t]
  \centering
  \includegraphics[width=1.0\linewidth]{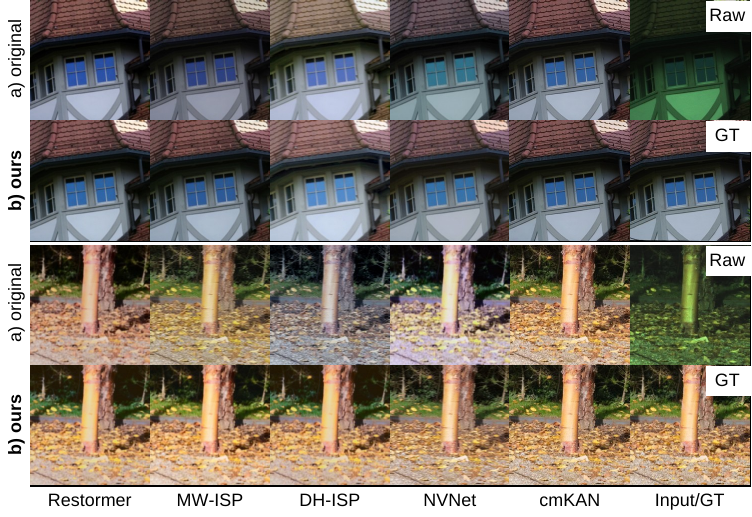}

   \caption{Quantitative results on the on the Mobile AI raw-to-sRGB dataset~\cite{ignatov2025learned}, demonstrating the architecture-agnostic nature of our EGUOT framework. We apply (a) the original, \textit{paired} training and (b) our proposed \textit{unpaired} training to a wide range of SOTA ISP backbones. Our unpaired mode achieves performance that is highly competitive with the original settings. Best viewed in the electronic version.}
   \label{fig:aim-isp}
\end{figure}

\begin{table}[t]
  \centering
    \caption{Results on the Mobile AI raw-to-sRGB dataset~\cite{ignatov2025learned}, comparing SOTA backbones retrained with our framework against their \textit{original paired-data results}. Our \textit{unpaired} mode achieves competitive performance. Better results are highlighted in yellow. \label{tab:mai-isp}}
  \resizebox{1.\linewidth}{!}
  {
  \begin{tabular}{@{}l|cccc|cccc@{}}
    \toprule
    Backbone & PSNR $\uparrow$ & SSIM $\uparrow$ & $\Delta E$ $\downarrow$  & LPIPS $\downarrow$ & PSNR $\uparrow$ & SSIM $\uparrow$ & $\Delta E$ $\downarrow$  & LPIPS $\downarrow$\\
    & \multicolumn{4}{c|}{\colorbox{supervised}{Paired Mode (original)}} & \multicolumn{4}{c}{\colorbox{unsipervised}{Unpaired Mode (ours)}}\\
    \midrule
    Restormer~\cite{zamir2022restormer}                    & 23.99 & 0.88 & 6.74 & 0.13 & 22.51 & 0.86 & 7.27 & 0.14 \\
    MW-ISP~\cite{ignatov2020aim}         & 24.31 & 0.88 & 6.27 & 0.11 & \colorbox{best}{24.44} & 0.87 & \colorbox{best}{6.09} & 0.11 \\
    AW-Net~\cite{dai2020awnet}                       & 24.17 & 0.87 & 6.35 & 0.12 & 24.06 & 0.87 & \colorbox{best}{6.33} & 0.12 \\
    LAN-ISP~\cite{wirzberger2022lan}     & 23.48 & 0.87 & 7.18 & 0.15 & 23.15 & 0.86 & 7.22 & \colorbox{best}{0.14} \\
    LiteISP~\cite{zhang2021learning}     & 24.05 & 0.86 & 6.43 & 0.13 & 24.01 & 0.86 & 6.58 & 0.13 \\
    DH-ISP~\cite{ignatov2021learned}     & 23.20 & 0.84 & 8.52 & 0.17 & 22.98 & 0.84 & \colorbox{best}{8.49} & 0.17 \\
    MicroISP~\cite{ignatov2022microisp}  & 23.87 & 0.85 & 7.04 & 0.16 & 23.79 & 0.85 & 7.19 & 0.16 \\
    NVNet~\cite{ignatov2025learned}                        & 24.09 & 0.85 & 6.19 & 0.13 & 23.87 & 0.84 & 7.03 & 0.13 \\
    cmKAN~\cite{nikonorov2025color}                        & 24.51 & 0.88 & 5.31 & 0.10 & 24.25 & 0.88 & 6.07 & 0.10 \\
    \bottomrule
  \end{tabular}
  } 
\end{table}

Finally, we validate our approach on the Mobile AI dataset~\cite{ignatov2025learned} in Table \ref{tab:mai-isp} and Figure \ref{fig:aim-isp}. The results are consistent with our findings on the other two datasets. Our unpaired mode remains highly competitive with paired settings, and in the case of MW-ISP~\cite{ignatov2020aim}, even outperforms the original model in both PSNR (24.44 vs 24.31) and $\Delta E$ (6.09 vs 6.27).

\subsection{Ablation Studies}

We conduct a series of ablation studies on the ZRR dataset~\cite{ignatov2020replacing} using the Restormer~\cite{zamir2022restormer} backbone to validate our core design choices.

\begin{figure}[t]
  \centering
  \includegraphics[width=1.0\linewidth]{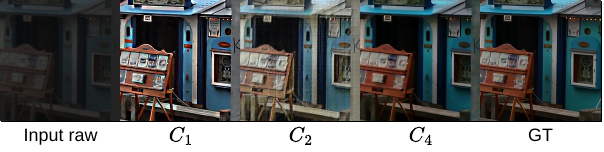}

   \caption{Visual ablation study on the ZRR dataset~\cite{ignatov2020replacing}. Our full method ($C_4$) produces the best results. The standard GAN baseline ($C_1$) suffers from artifacts and color issues. The UOT-Only baseline ($C_2$) fails to reconstruct fine textures and accurate color. The $\tau=0$ baseline ($C_3$) is not shown as it suffered from total mode collapse.}
   \label{fig:ablation-arch}
\end{figure}

\begin{table}[t]
  \centering
  \caption{
    Ablation study of our framework's core components on the ZRR~\cite{ignatov2020replacing} dataset, using a fixed Restormer~\cite{zamir2022restormer} backbone. Our full method ($C_4$) significantly outperforms a standard GAN baseline ($C_1$), while $C_2$ and $C_3$ demonstrate that both the expert committee and the UOT content cost ($\tau>0$) are essential for high-fidelity results.
  }
  \vspace{-2mm}
  \resizebox{\linewidth}{!}
  {
  \begin{tabular}{@{}llc|ccc@{}}
    \toprule
    Config & Adversarial Obj. & Expert Committee & PSNR $\uparrow$ & SSIM $\uparrow$ & $\Delta E \downarrow$ \\
    \midrule
    $C_1$ & Hinge Loss & \checkmark (All) & 19.23 & 0.73 & 10.8 \\
    $C_2$ & UOT-Loss & \textemdash & 19.15 & 0.72 & 12.2 \\
    $C_3$ & UOT-Loss ($\tau$=0) & \checkmark (All) & \multicolumn{3}{c}{Fails (Mode Collapse)} \\
    $C_4$ (\textbf{Ours}) & UOT-Loss & \checkmark (All) & \colorbox{best}{21.69} &  \colorbox{best}{0.84} &  \colorbox{best}{9.93} \\
    
    \bottomrule
  \end{tabular}
  }
  \vspace{-2mm}
  \label{tab:framework-ablation}
\end{table}

\vspace{-4mm}
\paragraph{Framework Ablation} We first validate our hybrid EGUOT framework in Table \ref{tab:framework-ablation}. We test four configurations:
($C_1$) a standard GAN baseline using Hinge Loss and our expert committee;
($C_2$) a "UOT-Only" baseline, using only the UOT objective without the expert committee;
($C_3$) our full model, but with the content cost weight $\tau$ set to 0;
and ($C_4$) our full EGUOT model.
The results clearly demonstrate that all components are essential. The GAN baseline ($C_1$) and the UOT-Only baseline ($C_2$) produce poor results, with low PSNR/SSIM and high color error. $C_3$ fails due to mode collapse, proving that the content cost $c(\cdot, \cdot)$ is essential to anchor the UOT mapping. Our full model ($C_4$) significantly outperforms all other configurations, showing that the UOT objective and the expert committee work synergistically.

\begin{figure}[t]
  \centering
  \includegraphics[width=0.8\linewidth]{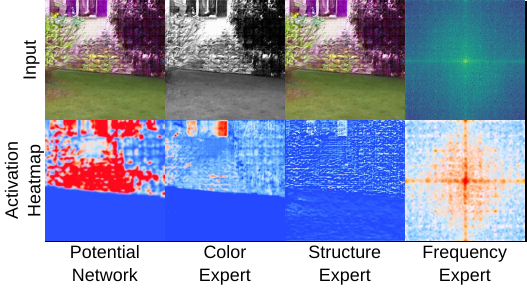}

   \caption{Activation heatmaps (red=fake, blue=real) showing our potential and the discriminator committee analyzing a fake sRGB image ($\hat{Y}$). Each network learns to spot specific errors: \textit{Potential Network} identifies global implausibility; \textit{Color Expert} highlights semantically incorrect colors; \textit{Structure Expert}Detects bad local textures; \textit{Frequency Expert}uses the FFT spectrum to find unnatural high-frequency artifacts (spikes).}
   \label{fig:ablation-maps}
\end{figure}

\begin{table}[t]
  \centering
  \caption{
    Ablation study of the expert committee. All models are trained with our UOT objective
    on the ZRR dataset~\cite{ignatov2020replacing} with the Restormer~\cite{zamir2022restormer} backbone.
    The baseline ($C_2$) uses no expert discriminators.
    Our full method ($C_4$) utilizes all experts, and removing any of them,
    especially the Color Expert ($E_1$), leads to a significant drop in performance.
  }
  \vspace{-2mm}
  \resizebox{\linewidth}{!}
  {
  \begin{tabular}{@{}lccc | ccc @{}}
    \toprule
    Config & Color Exp. & Structure Exp. & Frequency Exp. & PSNR $\uparrow$ & SSIM $\uparrow$ & $\Delta E \downarrow$ \\
    \midrule
    
    $C_2$ (UOT-Only)      & \textemdash & \textemdash & \textemdash & 19.15 & 0.72 & 12.2 \\
    \midrule

    $E_1$ (w/o Color)     & \textemdash & \checkmark & \checkmark & 19.88 & 0.75 & 14.8 \\
    $E_2$ (w/o Structure) & \checkmark & \textemdash & \checkmark & 20.16 & 0.79 & 10.4 \\
    $E_3$ (w/o Frequency) & \checkmark & \checkmark & \textemdash & 20.95 & 0.82 & 9.91 \\

    $C_3$ (\textbf{Ours}) & \textbf{\checkmark} & \textbf{\checkmark} & \textbf{\checkmark} & \colorbox{best}{21.69} &  \colorbox{best}{0.84} &  \colorbox{best}{9.93} \\
    
    \bottomrule
  \end{tabular}
  }
  \vspace{-2mm}
  \label{tab:expert-ablation}
\end{table}

\vspace{-4mm}
\paragraph{Expert Committee Ablation} In Table \ref{tab:expert-ablation}, we analyze the contribution of each expert discriminator. We start from the $C_2$ (UOT-Only) baseline and also show the results of removing one expert at a time from our full model ($C_4$). The results show that all three experts contribute to the final performance. The UOT-Only model ($C_2$) is the weakest. Removing any single expert degrades performance, with the Color Expert being the most critical ($E_1$); its removal causes the $\Delta E$ to spike from 9.93 to 14.8, confirming its vital role in maintaining color fidelity. Figure \ref{fig:ablation-maps} provides a qualitative visualization of this specialization, showing how each component learns to identify distinct error types.

\section{Conclusion}
\label{sec:conclusion}

In this work, we addressed the critical bottleneck in modern ISP development: the reliance on large-scale, paired raw-sRGB datasets. We identified that prior unpaired methods often fail due to their sensitivity to the distributional outliers found in real-world target sRGB datasets.

To solve this, we proposed \textbf{EGUOT}, a novel and robust framework for raw-to-sRGB translation, designed to operate in both paired and unpaired modes. Our framework is the first to apply Unbalanced Optimal Transport (UOT) for the ISP task, leveraging its theoretical robustness to dataset outliers. To achieve the high fidelity required of a modern ISP, we introduced a Committee of Expert Discriminators. This hybrid regularizer guides the mapping by providing targeted perceptual gradients for color, structure, and frequency, and is a key component in both training modes.

Our comprehensive experiments demonstrated the power and flexibility of our approach. We showed that EGUOT is architecture-agnostic, capable of training numerous state-of-the-art ISPs in a fully unpaired manner. Our paired training mode consistently exceeds the performance of the original settings across all metrics. Concurrently, our unpaired results are highly competitive with, and in some cases surpass, their original fully-paired counterparts. Furthermore, our unpaired mode sets a new state-of-the-art for unpaired raw-to-sRGB translation, significantly outperforming all prior GAN and OT-based methods. Our ablation studies empirically confirmed our core hypotheses: the UOT objective is essential for outlier robustness, while the expert committee is critical for perceptual fidelity.

We believe this approach will make ISP development more efficient by allowing developers to focus on the desired distribution properties instead of spending extensive time on data collection.

\subsubsection*{Limitations and Future Work}
While our expert committee is effective, it adds complexity compared to a single-discriminator framework. Future work could explore methods to prune this committee. Moreover, the demonstrated effectiveness of EGUOT opens promising avenues for applying this framework to other challenging unpaired image restoration tasks, such as low-light enhancement, denoising, and HDR reconstruction, where data outliers are common.
\section*{Acknowledgments}

This work was partly supported by The Alexander von Humboldt Foundation.
The authors also express sincere gratitude to Alexander Korotin for his valuable advice on the theoretical background of Optimal Transport. 

{
    \small
    \bibliographystyle{ieeenat_fullname}
    \bibliography{main}
}

\clearpage
\setcounter{page}{1}
\maketitlesupplementary

\textit{
In this supplementary material, we first present additional quantitative and qualitative results in Sec.~\ref{sec:extra-results} to further validate the architecture-agnostic nature of our framework. Next, we provide a detailed robustness study in Sec.~\ref{sec:robustness}, empirically demonstrating the superiority of Unbalanced Optimal Transport over standard GANs in the presence of dataset outliers.
}

\section{Additional results}
\label{sec:extra-results}

In this section, we verify the performance of the proposed EGUOT framework across three diverse benchmarks: the Zurich Raw-to-RGB (ZRR) dataset~\cite{ignatov2020replacing}, the Mobile AI (MAI) dataset~\cite{ignatov2025learned}, and the ISPIW dataset~\cite{shekhar2022transform}.

\begin{figure}[h]
  \centering
  \includegraphics[width=1.0\linewidth]{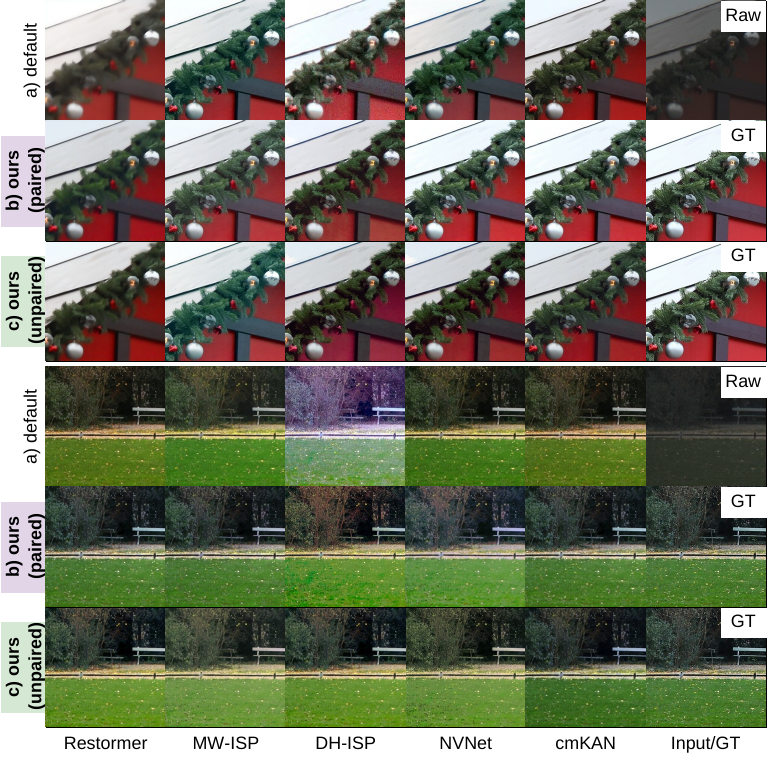}

   \caption{Quantitative results on the on the ZRR raw-to-sRGB dataset~\cite{ignatov2020replacing}, demonstrating the architecture-agnostic nature of our EGUOT framework. We apply (a) the original, \textit{paired} training and our proposed (b) \textit{paired} / (c) \textit{unpaired} training to a wide range of SOTA ISP backbones. Our \textit{paired} mode consistently outperforms the original, while our \textit{unpaired} mode achieves competitive performance. Best viewed in the electronic version.}
   \label{fig:zrr-isp2}
\end{figure}

\begin{figure}[h]
  \centering
  \includegraphics[width=1.0\linewidth]{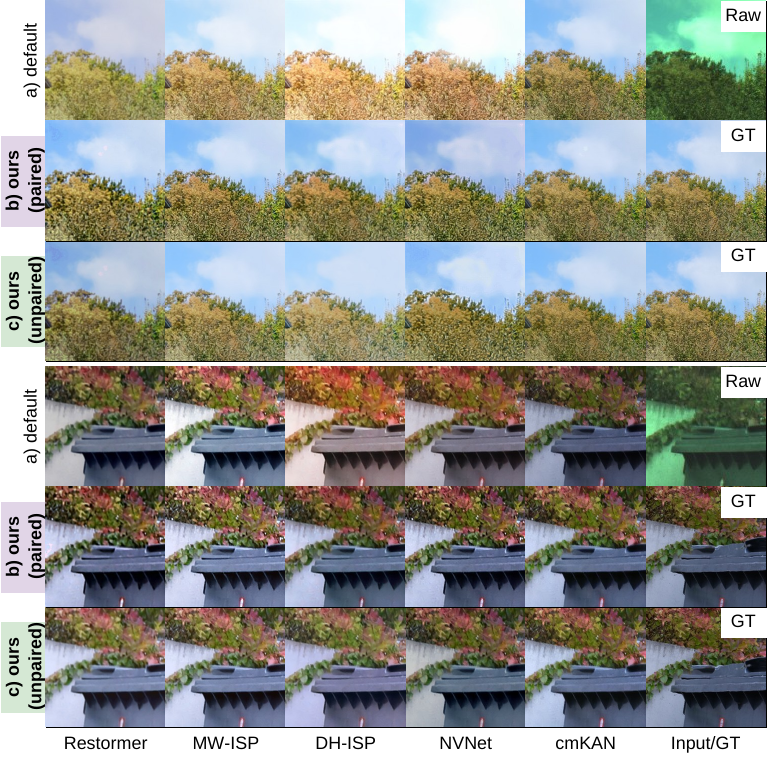}

   \caption{Quantitative results on the on the MAI raw-to-sRGB dataset~\cite{ignatov2025learned}, demonstrating the architecture-agnostic nature of our EGUOT framework. We apply (a) the original, \textit{paired} training and our proposed (b) \textit{paired} / (c) \textit{unpaired} training to a wide range of SOTA ISP backbones. Our \textit{paired} mode consistently outperforms the original, while our \textit{unpaired} mode achieves competitive performance. Best viewed in the electronic version.}
   \label{fig:mai-isp2}
\end{figure}

\begin{figure}[h]
  \centering
  \includegraphics[width=1.0\linewidth]{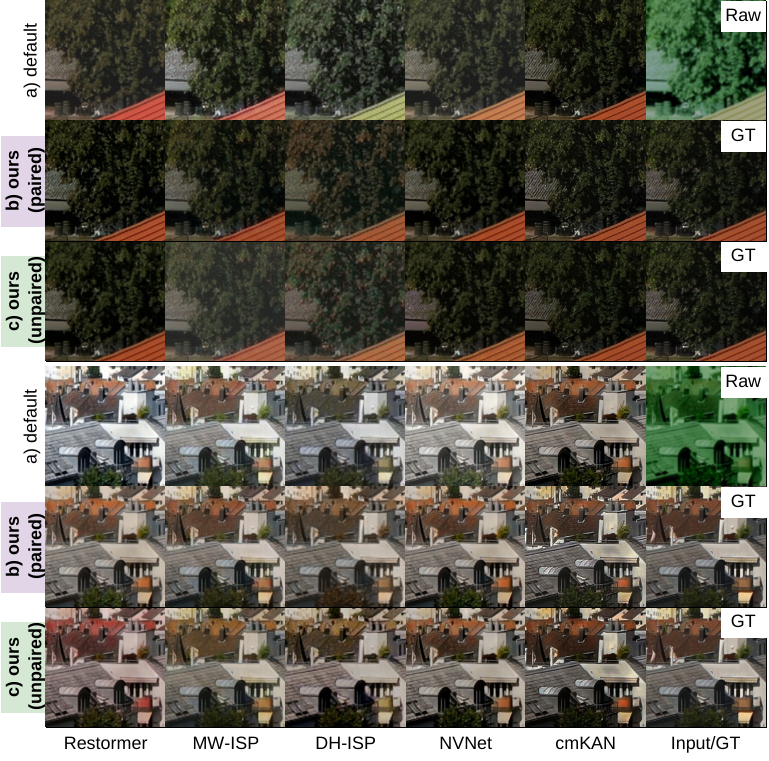}

   \caption{Quantitative results on the on the ISPIW raw-to-sRGB dataset~\cite{shekhar2022transform}, demonstrating the architecture-agnostic nature of our EGUOT framework. We apply (a) the original, \textit{paired} training and our proposed (b) \textit{paired} / (c) \textit{unpaired} training to a wide range of SOTA ISP backbones. Our \textit{paired} mode consistently outperforms the original, while our \textit{unpaired} mode achieves competitive performance. Best viewed in the electronic version.}
   \label{fig:ispiw-isp2}
\end{figure}

\textbf{Visual Quality.} We provide extensive visual comparisons in Fig.~\ref{fig:zrr-isp2}, Fig.~\ref{fig:mai-isp2}, and Fig.~\ref{fig:ispiw-isp2}. These qualitative results demonstrate that our Experts-Guided Unbalanced Optimal Transport (EGUOT) framework successfully reconstructs high-fidelity sRGB images across different sensors and scenes. Crucially, our \textit{unpaired} training mode produces results that are perceptually indistinguishable from, and in some cases sharper than, the original \textit{paired} baselines. The Expert Committee—specifically, the frequency and structure experts—ensures that the generated images remain free of the hallucinatory artifacts often observed in standard GAN-based unpaired translations.

\textbf{Quantitative Evaluation.} We report detailed quantitative metrics for our paired training mode in Table~\ref{tab:mai-isp2} (Mobile AI) and Table~\ref{tab:ispiw-isp2} (ISPIW). Consistent with the results reported in the main paper, our framework establishes a new state-of-the-art for supervised training. By utilizing the Expert Committee as a regularizer alongside the pixel-wise loss, we consistently improve upon the original paired baselines across all metrics (PSNR, SSIM, $\Delta E$, and LPIPS). 

For instance, on the Mobile AI dataset (Table~\ref{tab:mai-isp2}), our method improves the MW-ISP~\cite{ignatov2020aim} backbone by nearly 0.5 dB in PSNR while reducing the color error ($\Delta E$) from 6.27 to 5.91. Similarly, on the ISPIW dataset (Table~\ref{tab:ispiw-isp2}), our training scheme boosts the performance of the recent Transformer-based Restormer~\cite{zamir2022restormer} and LiteISP~\cite{zhang2021learning} architectures. These results confirm that EGUOT is a robust, architecture-agnostic training paradigm that enhances ISP learning regardless of the underlying network capacity.

\begin{table}[h]
  \centering
    \caption{Results on the Mobile AI raw-to-sRGB dataset~\cite{ignatov2025learned}, comparing SOTA backbones retrained with our framework against their \textit{original paired-data results}. Our \textit{paired} mode achieves the best performance. Better results are highlighted in yellow. \label{tab:mai-isp2}}
  \resizebox{1.\linewidth}{!}
  {
  \begin{tabular}{@{}l|cccc|cccc@{}}
    \toprule
    Backbone & PSNR $\uparrow$ & SSIM $\uparrow$ & $\Delta E$ $\downarrow$  & LPIPS $\downarrow$ & PSNR $\uparrow$ & SSIM $\uparrow$ & $\Delta E$ $\downarrow$  & LPIPS $\downarrow$\\
    & \multicolumn{4}{c|}{\colorbox{supervised}{Paired Mode (original)}} & \multicolumn{4}{c}{\colorbox{ourssuper}{Paired Mode (ours)}}\\
    \midrule
    Restormer~\cite{zamir2022restormer}  & 23.99 & 0.88 & 6.74 & 0.13 & \colorbox{best}{24.09} & 0.88 & \colorbox{best}{6.67} & \colorbox{best}{0.12} \\
    MW-ISP~\cite{ignatov2020aim}         & 24.31 & 0.88 & 6.27 & 0.11 & \colorbox{best}{24.79} & \colorbox{best}{0.89} & \colorbox{best}{5.91} & \colorbox{best}{0.10} \\
    AW-Net~\cite{dai2020awnet}           & 24.17 & 0.87 & 6.35 & 0.12 & \colorbox{best}{24.33} & \colorbox{best}{0.88} & \colorbox{best}{6.12} & \colorbox{best}{0.11} \\
    LAN-ISP~\cite{wirzberger2022lan}     & 23.48 & 0.87 & 7.18 & 0.15 & \colorbox{best}{23.59} & \colorbox{best}{0.88} & \colorbox{best}{7.07} & \colorbox{best}{0.14} \\
    LiteISP~\cite{zhang2021learning}     & 24.05 & 0.86 & 6.43 & 0.13 & \colorbox{best}{24.22} & \colorbox{best}{0.87} & \colorbox{best}{6.08} & \colorbox{best}{0.12} \\
    DH-ISP~\cite{ignatov2021learned}     & 23.20 & 0.84 & 8.52 & 0.17 & \colorbox{best}{23.29} & 0.84 & \colorbox{best}{8.21} & \colorbox{best}{0.16} \\
    MicroISP~\cite{ignatov2022microisp}  & 23.87 & 0.85 & 7.04 & 0.16 & \colorbox{best}{24.00} & \colorbox{best}{0.86} & \colorbox{best}{6.99} & \colorbox{best}{0.15} \\
    NVNet~\cite{ignatov2025learned}      & 24.09 & 0.85 & 6.19 & 0.13 & \colorbox{best}{24.12} & \colorbox{best}{0.86} & \colorbox{best}{6.11} & 0.13 \\
    cmKAN~\cite{nikonorov2025color}      & 24.51 & 0.88 & 5.31 & 0.10 & \colorbox{best}{24.55} & 0.88 & \colorbox{best}{5.28} & 0.10 \\
    \bottomrule
  \end{tabular}
  } 
\end{table}

\begin{table}[h]
  \centering
    \caption{Results on the ISPIW raw-to-sRGB dataset~\cite{shekhar2022transform}, comparing SOTA backbones retrained with our framework against their \textit{original paired-data results}. Our \textit{paired} mode achieves the best performance. Better results are highlighted in yellow.\label{tab:ispiw-isp2}}
  \resizebox{1.\linewidth}{!}
  {
  \begin{tabular}{@{}l|cccc|cccc@{}}
    \toprule
    Backbone & PSNR $\uparrow$ & SSIM $\uparrow$ & $\Delta E$ $\downarrow$  & LPIPS $\downarrow$ & PSNR $\uparrow$ & SSIM $\uparrow$ & $\Delta E$ $\downarrow$  & LPIPS $\downarrow$\\
    & \multicolumn{4}{c|}{\colorbox{supervised}{Paired Mode (original)}} & \multicolumn{4}{c}{\colorbox{ourssuper}{Paired Mode (ours)}}\\
    \midrule
    Restormer~\cite{zamir2022restormer} & 20.93 & 0.79 & 6.85 & 0.14 & \colorbox{best}{21.04} & \colorbox{best}{0.80} & \colorbox{best}{6.78} & \colorbox{best}{0.13} \\
    MW-ISP~\cite{ignatov2020aim}        & 21.90 & 0.81 & 7.03 & 0.11 & \colorbox{best}{21.96} & 0.81 & \colorbox{best}{6.85} & \colorbox{best}{0.10} \\
    AW-Net~\cite{dai2020awnet}          & 21.75 & 0.81 & 6.99 & 0.12 & \colorbox{best}{22.04} & \colorbox{best}{0.82} & \colorbox{best}{6.87} & \colorbox{best}{0.11} \\
    LAN-ISP~\cite{wirzberger2022lan}    & 22.09 & 0.81 & 6.97 & 0.11 & \colorbox{best}{22.11} & \colorbox{best}{0.82} & \colorbox{best}{6.40} & \colorbox{best}{0.10} \\
    LiteISP~\cite{zhang2021learning}    & 22.14 & 0.81 & 6.31 & 0.11 & \colorbox{best}{22.33} & \colorbox{best}{0.82} & \colorbox{best}{6.09} & \colorbox{best}{0.10} \\
    DH-ISP~\cite{ignatov2021learned}    & 19.92 & 0.76 & 7.89 & 0.17 & \colorbox{best}{21.15} & \colorbox{best}{0.77} & \colorbox{best}{7.27} & 0.17 \\
    MicroISP~\cite{ignatov2022microisp} & 20.70 & 0.77 & 6.92 & 0.15 & \colorbox{best}{20.88} & \colorbox{best}{0.78} & \colorbox{best}{6.29} & \colorbox{best}{0.14} \\
    NVNet~\cite{ignatov2025learned}     & 23.80 & 0.82 & 5.81 & 0.10 & \colorbox{best}{23.96} & \colorbox{best}{0.83} & \colorbox{best}{5.80} & 0.10 \\
    cmKAN~\cite{nikonorov2025color}     & 24.22 & 0.83 & 5.29 & 0.09 & \colorbox{best}{24.30} & \colorbox{best}{0.84} & \colorbox{best}{5.15} & 0.09 \\
    \bottomrule
  \end{tabular}
  }
\end{table}

\section{Robustness Study}
\label{sec:robustness}

\begin{table}[h]
  \centering
  \caption{
    Robustness to Outliers. We compare our UOT-based framework against a standard GAN
    when trained on a ``clean'' vs. ``dirty'' (15\% outliers) ZRR dataset~\cite{ignatov2020replacing}. 
    Our method's performance remains stable, demonstrating its robustness.
  }
  \vspace{-2mm}
  \resizebox{0.7\linewidth}{!}
  {
  \begin{tabular}{@{}ll|ccc@{}}
    \toprule
    Config & Training Dataset & PSNR $\uparrow$ & SSIM $\uparrow$ & $\Delta E \downarrow$ \\
    \midrule
    \multirow{2}{*}{$C_1$} & Clean & 19.23 & 0.73 & 10.8 \\
     & \textbf{Dirty} & \textbf{18.65} & \textbf{0.72} & \textbf{11.17} \\
    \midrule
    \multirow{2}{*}{$C_4$ (\textbf{Ours})} & Clean & \colorbox{best}{21.69} & \colorbox{best}{0.84} & \colorbox{best}{9.93} \\
     & \textbf{Dirty} & \colorbox{best}{\textbf{21.41}} & \colorbox{best}{\textbf{0.84}} & \colorbox{best}{\textbf{10.09}} \\
    \bottomrule
  \end{tabular}
  }
  \vspace{-2mm}
  \label{tab:robustness-ablation}
\end{table}

A core motivation of our work is the fragility of standard GAN losses when facing noisy datasets—a common scenario in real-world unpaired learning where the source and target domains contain different outliers. To validate our hypothesis that the Unbalanced Optimal Transport (UOT) framework provides superior robustness, we conduct a controlled corruption experiment.

We create a ``Dirty'' version of the ZRR~\cite{ignatov2020replacing} target dataset by polluting it with 15\% outlier images. Specifically, we apply severe random augmentations (random color jitter, contrast shifts, and brightness modifications) to 15\% of the training data to simulate low-quality samples often found in scraped datasets. We then train our full model ($C_4$) and a standard GAN baseline ($C_1$, utilizing a Hinge loss) on both the ``Clean'' and ``Dirty'' datasets.

As shown in Table~\ref{tab:robustness-ablation}, the standard GAN's performance is highly sensitive to the data quality. When trained on the polluted data, the GAN fails to discount the outliers, leading to a significant performance degradation: its PSNR drops by 0.58 dB and $\Delta E$ worsens by 0.37 units. In stark contrast, our EGUOT framework demonstrates remarkable stability. Consequently, our method maintains high fidelity on the ``Dirty'' dataset, suffering only a negligible 0.28 dB drop in PSNR and 0.18 units in $\Delta E$. This empirically proves that UOT-based framework is fundamentally more robust to the noisy data distributions encountered in unpaired environments.

\end{document}